# Probabilistic Reasoning About Ship Images

*Lashon B. Booker*

Navy Center for Applied Research in AI
Code 7510
Naval Research Laboratory
Washington, D.C. 20375

*Naveen Hota*

JAYCOR
1608 Spring Hill Road
Vienna, VA 22180

## ABSTRACT

*One of the most important aspects of current expert systems technology is the ability to make causal inferences about the impact of new evidence. When the domain knowledge and problem knowledge are uncertain and incomplete, Bayesian reasoning has proven to be an effective way of forming such inferences [3,4,8]. While several reasoning schemes have been developed based on Bayes Rule, there has been very little work examining the comparative effectiveness of these schemes in a real application. This paper describes a knowledge based system for ship classification [1], originally developed using the PROSPECTOR updating method [2], that has been reimplemented to use the inference procedure developed by Pearl and Kim [4,5]. We discuss our reasons for making this change, the implementation of the new inference engine, and the comparative performance of the two versions of the system.*

## INTRODUCTION

Classifying images is extremely difficult whenever the feature information available is incomplete or uncertain. Under such circumstances, identification of an object requires some kind of reasoning mechanism to help resolve ambiguous interpretations within the constraints of the available domain knowledge. The need for a reasoning mechanism becomes even more acute if the interpretation process is also constrained by limited resources. When there is not enough time or memory for an exhaustive feature analysis, intelligent decisions must be made about how to use the resources available to maximum advantage. This means that the reasoning mechanism must be involved in the control of the information extraction activities, as well as the interpretation of the results. Ship classification is an example of one practical application in which all of these problems arise.

Classification of ships in an operational environment is a difficult task regardless of what kind of images are used. This is not always obvious to those who are only familiar with the detailed views of a ship one finds in a reference book. Observers in the field rarely have the luxury of an abundance of clear details to work with. Images are most often obtained during a *brief* observation interval from a distance that makes high resolution difficult to achieve. The viewing angle is usually a matter of opportunity rather than choice, and the observer must make do with the prevailing visibility, weather, and lighting conditions at sea. Another factor degrading image quality is

29

the fact that sensor platforms are often buffeted by turbulence in the air or the ocean. The quality of images produced in this way is likely to be lower than that attainable using sophisticated enhancement techniques and powerful computing resources. These difficulties are of course exacerbated when the classification must be done in real time. All of this is in addition to the complexity faced when distinguishing among hundreds of classes of vessels, some of which differ only in fine feature details.

Having this task performed well is obviously important to the Navy, which has invested heavily in training personnel to analyze and interpret images under operational conditions. The human observer-sensor operator must be highly trained and experienced. He/she must know which features are related to which ship classes, and make a judgement as to how well various features are manifested in the image. Moreover, the observer must keep track of the implications of all these judgements — both with respect to their uncertainty and consistency, and with respect to an eventual classification. A decision aid must also cope with these problems, but in a way that acknowledges the meager computational resources available on most military platforms — an important constraint now and in the near future. The most useful kind of system is one that can distinguish *similar* ship types. Most trained personnel can easily tell the difference between an aircraft carrier and a cruiser. It is much more difficult to make decisions about several types of cruisers whose images are similar.

Real-time ship classification is a demanding application. It is a task requiring that complex inferences, based on incomplete and uncertain information, be made reliably under stringent computational constraints. In devising a system that meets this challenge, two of the most important research issues are control and inference. Given that time constraints often preclude an exhaustive feature analysis, which features should be sought after in the time available? Given an uncertain and incomplete feature description, what kind of heuristic reasoning tools provide reliable and computationally inexpensive ship classifications? This paper describes a knowledge-based system for reasoning about ship images that successfully manages many of these issues. A prototype developed at the Navy Center for Applied Research in AI (NCARAI) has convincingly demonstrated that a heuristic approach to this problem is effective and practical. Our current research effort builds on this work, and is developing a 2nd generation expert system to help solve this classification problem.

## REASONING ABOUT SHIP IMAGES

The focus of this research is on how to use incomplete and uncertain feature information to make plausible inferences about Naval Class.*

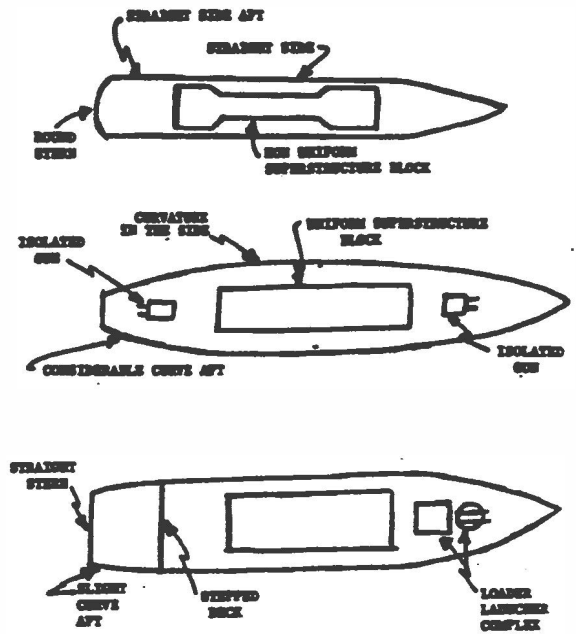

Figure 1 Examples of plan view features

Reasoning about plausible classifications for a ship image requires knowledge about the features needed to describe various Naval Classes; and, knowledge about how the presence or absence of these features in an image implies one class versus another. Feature details might be observable from either a *profile* (or side) view, a *plan* (or top down) view, or both. Figure 1 shows the kinds of features that are important in analyzing a plan view image. The primary items of interest are the shape of the stern, curvature of the sides, superstructure configuration, etc. Needless to say, not all of this detail is likely to be available in every image, and an analyst often has to make uncertain judgements about whether or not they are really there.

---

* A Naval Class is a group of ships built to the same design and known by the lead ship's name.

This knowledge can be organized into a simple hierarchy having four levels: Naval Class, major structural components, features, and observations. At the top of the hierarchy are the hypotheses about how to classify a particular image. This presumably cannot be directly determined, so at the next level are hypotheses about the gross structural components of the ship — the stern, deck, superstructure blocks, etc. Sometimes evidence is available that directly bears on knowledge at this level. For instance, the stern of the Sverdlov class is very distinctive and can often be recognized immediately. In most cases, though, components have to be determined from lower level attributes. These lower level structural attributes, in turn, are usually established based on knowledge about what is manifested in the imagery.

### TABLE 1
### PLAN VIEW STERN DESCRIPTIONS

| Stern Type | Shape Attribute | | |
|---|---|---|---|
| | Square | Round | Tapered |
| Virginia | 10 | 0 | 0 |
| Belknap Leahy | 0 | 10 | 0 |
| Sverdlov | 1 | 0 | 10 |
| Bainbridge California Coontz Long Beach Truxtun | 0 | 5 | 0 |
| Forrest Sherman | 1 | 2 | 0 |

More specifically, consider the following example from a real Navy problem. Table 1 summarizes an expert analyst's description of the stern component for plan views of 10 classes. There is an implicit knowledge hierarchy in this description. At the component level, 5 types of stern components are represented here. The description of each type includes a subjective weight for each shape attribute. This number indicates an expectation about whether that attribute will be manifested in the imagery. The weights are given on a scale of 0 to 10, with 0 meaning the attribute should never be detected and 10 meaning it should always be detected. Two structures with the same weight for a given attribute cannot be distinguished on that basis alone. So, for example, the sterns of Sverdlov and Forrest Sherman are square in the same way. Such knowledge about the specific nature of the attributes resides at the level directly below the component level in the hierarchy. For this set of ships, there are two ways for a stern to be square, three ways to be round, and one way to be tapered. In practice, it is often difficult even to obtain evidence at this level. A poor quality image or non-expert observer might only be able to provide evidence that the stern in the image is somewhat rounded, period. The lowest level in the hierarchy represents these very simple assertions.

The hierarchical organization of knowledge corresponds to the reasoning steps involved in classifying an image: image analysis or user observations provide data about features; feature information can be used to infer higher order ship components; and, the higher-order components provide a much simpler basis for determining Naval Class. The process does not proceed in only a "bottom-up" fashion however. At each stage, the features extracted so far designate a set of likely candidate classes. Each such class can be used as a source of prediction about additional features that should be in the image, thus directing the feature extraction process in a "top-down" manner. One AI framework made to order for this hierarchical reasoning task is the *inference network*, which directly represents the causal influences among propositions or variables of interest. Each node in such a network represents a proposition (or variable) describing some aspect of the domain. Each link signifies a direct dependency between two

31

propositions. The network not only can be used as a knowledge structure in which facts about the domain are stored, but it can also provide a computational framework for reasoning about that knowledge. If some measure of belief is associated with each node, and the dependencies between nodes are summarized by constraints on beliefs, then the network structure indicates which beliefs need to be updated when new information is available. In this way, an integrated summary can be maintained of what is known directly about each hypothesis and what can be inferred from their inter-relationships.

## A SIMPLE PROTOTYPE

In order to demonstrate that the inference network approach is an effective way to deal with the ship classification problem, a prototype decision aid was developed and tested [1]. The knowledge for this system was provided by an expert analyst who picked 10 Naval Classes that have similar imagery and are often difficult to distinguish from one another. Feature descriptions for the plan and profile views of each class were given in the manner shown in Table 1. Because the number of ships belonging to each class is known in advance, simple counting arguments can be used to quantify the relative beliefs and constraints on beliefs associated with this knowledge. Probabilities are therefore a very natural measure of belief to use in the system. Several probabilistic reasoning schemes have been devised for updating beliefs in inference networks [8]. A version of the PROSPECTOR updating method [2] was developed at NCARAI to solve a resource allocation problem [10], and was available for the ship classification work. Consequently, a PROSPECTOR-style inference engine was used to implement the prototype.

In our version of the PROSPECTOR scheme, the relation between evidence and hypothesis is a rule of inference of the following type:

If P1 then (to extent $\lambda 1$, $\lambda 2$) conclude P.

P1 and P are both propositions. P1 is the antecedent of the rule and P is the consequent. The strength of the implication is attenuated by the two numbers $\lambda 1$ and $\lambda 2$: $\lambda 1$ is the conditional probability of P given that P1 is true; and, $\lambda 2$ is the conditional probability of P given that P1 is false. This information, together with the prior probability of P and P1, is used to compute a posterior probability for P when the truth of P1 is uncertain. When several independent propositions have an evidential relationship with P, the posterior probability of P is computed using a heuristic generalization of Bayes rule. See Duda et al. [2] for more details.† Reasoning is accomplished in this framework by *propagating* changes through the inference network. A change in the probability of a proposition causes the probabilities of its consequents to be updated as described above. The procedure is then recursively applied starting at each consequent. In this way, the effects of the initial change spread throughout the network to all propositions that are directly or indirectly related.

As a simplification, the prototype system was implemented to rely exclusively on operator input. This allows the reasoning issues to be examined without having to worry at all about feature extraction issues. The system interacts with the operator to get the feature information in a mixed-initiative fashion. At any point during the session the operator can volunteer information about the presence or absence of certain features in the image. In this way, the operator can direct the program's chain of reasoning in a manner he deems appropriate. When the operator is not volunteering information, the program asks a series of questions about the image. The questioning sequence is dynamically ordered so as to maximize the effectiveness of the evidence in determining a classification.

A global control strategy is used to select which question to ask. Each proposition is assigned a weight — called a *merit value* — proportional to its ability to alter the value of a top-level proposition. More specifically, the merit of a proposition H is the ratio $\delta P/\delta C$ where $\delta P$ is the expected change in the value of the top-level proposition if a value for H is obtained; and, $\delta C$ is

---

† The scheme also allows belief to be inferred between propositions related by logical AND, OR and NOT. However, only evidential relations were needed for the classification prototype.

32

the expected cost of obtaining a value for H. With this information, an efficient algorithm can be used that finds the proposition with the highest merit* in a network. See Slagle [9] for a complete discussion of how merit values are dynamically calculated and updated. Details about how these ideas were implemented in the prototype can be found in [1].

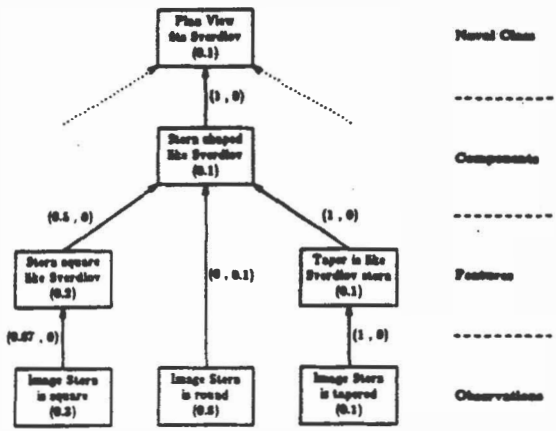

Figure 2. An excerpt from the hierarchical network used to determine how well a plan view image fits the Sverdlov class. Each rectangle designates a proposition along with its prior probability (the number in parentheses). The conditional probabilities λ1 and λ2 needed to do evidential updating are given as an ordered pair adjacent to each arc.

The structure of the inference networks derived from the feature descriptions is illustrated in Figure 2. This excerpt is a portion of the network for evaluating the hypothesis that a plan view image belongs to the Sverdlov class. It shows how the shape of the stern influences the top-level hypothesis. The bottom nodes correspond to the three relevant observations about the shape of the stern in the image. There are two propositions at the feature level, corresponding to the fact that the Sverdlov class is described as both square and tapered in Table 1. At the third level in the hierarchy, belief is computed about whether the overall stern shape fits Sverdlov. An observation that the stern is round is evidence that the stern does not fit Sverdlov, so its influence comes in at this point. It is also at this level that the feature weights are explicitly factored into the com-putation. The impact of evidence not sure to be detected (ie. with a weight less than 10) is modeled by changing the link parameters by the ratio weight/10. Networks of this type were constructed for all 10 Naval Classes, one set for plan views and another set for profiles. Overall, they contained 598 propositions and over 1000 links.

---
* Since merit is a signed quantity, what is intended here is the merit with highest absolute value.

The prototype system has been extensively tested on 101 images of the 10 classes. These images were photographs of sensor data from various sources, chosen because they are typical of the mediocre quality available from most operational systems. In 85 of the 101 trials, the ship class ranked first by the program was the correct classification. For profile images, the correct ship class was never ranked lower than second. The correct ship class was not as easily singled out for plan views. This is to be expected, however, given the relatively small amount of information available in a plan view image. Overall, the prototype reliably ranked the correct class at or near the top of the list. Navy experts have reviewed these results and judged them to be excellent given the quality of the test images.

## SCALING UP TO MORE REALISTIC PROBLEMS

Because of the interest generated by the performance of the prototype, work on this problem has now moved into a second phase in which a more realistic system is being developed. A system suitable for the operational Navy must have the capacity to represent and reason about any of the approximately 640 military Naval Classes that might be encountered at sea. The information available for making classification decisions will come from several sources. In many situations, human judgement and pattern recognition capabilities will continue to be an important source of information. Increasingly, however, information is becoming available from machine-generated feature analysis of raw sensor signals and imagery. Eventually, much of the classification process will be completely automated.

The research issues associated with this larger problem are considerably more complex than those dealt with in the feasibility study. For example, the knowledge base used in the prototype is much too shallow. It

33

is clear that more extensive feature descriptions will be needed to resolve ambiguities in a larger set of ships. More important, though, is the fact that more *kinds* of knowledge will be needed. At the very least, the system will have to know something about the reliability of the sensor being used, the physical relationships among ship components, and the many taxonomic relations among concepts related to the structure and function of ships. Knowledge about the classification process itself would also be useful, so that classification decisions can be made at a level of specificity commensurate with prevailing resource constraints. Moreover, an interface with signal and image processing modules sometimes requires the capability to represent knowledge about continuous-valued variables.

The increased complexity of the problem also has implications for the kinds of reasoning that will be necessary. Effective interaction with feature extraction modules will involve decisions about the order to acquire data, the number of image frames to process before making a judgement about some feature, etc. This means that inferences must flow from hypotheses to evidence as well as from evidence to hypotheses. Non-causal inferences will also be required. For instance, geometric reasoning about aspect angle and hidden features is extremely important. Clearly, the causal inference mechanism will have to interact smoothly with the other methods used for inference and control.

Taking all of these requirements into account, the PROSPECTOR inference scheme does not appear to be well suited for the larger problem. It only provides for data-driven inferences and works best with data driven control strategies. Because of the stringent independence assumptions made in this framework, sets of mutually exclusive and exhaustive multi-valued variables cannot be adequately modeled [3]. Consequently, only true-false propositions were used in the prototype system and the networks only encoded a selected subset of the dependencies among propositions. Even with these simplifications, however, the networks were difficult to maintain. The entire system used 2726 inter-related probabilities. As the domain expert refined the feature descriptions, managing the changes in so many probabilities became extremely difficult. A spreadsheet calculator database was constructed to alleviate some of the computational burden. An annoying conceptual burden still remained though. The network structure simply did not correspond closely enough to the intuitive picture of how the evidence interacts. These difficulties would of course be compounded in a larger, more complex classification problem.

## THE BMS APPROACH

After examining several alternatives, we have chosen the Belief Maintenance System (BMS) developed by Pearl and Kim [4,5] as the point of departure for this work. The BMS approach has several properties that fit nicely with the requirements of the ship classification problem:

- Both goal driven and data driven inferences are allowed.
- Updating is done with local computations that are independent of the control mechanism that initiates the process.
- Network nodes can represent discrete or continuous valued variables [6].
- A related mechanism can be used to maintain beliefs in object/class hierarchies [7].

These properties, together with the fact that beliefs are updated in a manner consistent with the axioms of probability theory, make BMS a good choice for this application.

The BMS procedure is a Bayesian updating scheme that keeps track of two sources of support for belief at each node: the diagnostic support of the data gathered by descendants of the node and the causal support of the data gathered by ancestors of the node. Each source of support is summarized by a separate local parameter. These two parameters, together with a matrix of conditional probabilities relating the node to its parents, are all that is required to update beliefs. Incoming evidence perturbates one or both of the support parameters for a node. This serves as an activation signal, causing belief at that node to be recomputed and support for neighboring nodes to be revised. The revised support is transmitted to the neighboring nodes, thereby propagating the impact of the evidence. Propagation continues until the network reaches equilibrium. See |4| for a more detailed description.

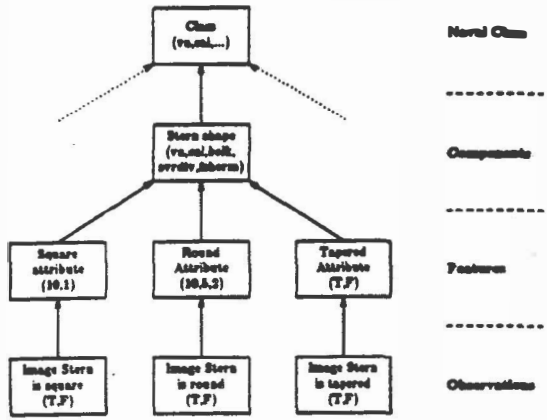

Figure 3 An excerpt from the BMS network used to determine how to classify a plan view image. Each rectangle designates a variable along with its mutually exclusive and exhaustive values (given in parentheses).

We have completed an object-oriented implementation of this procedure and tested it on the problem formulated for the original classification prototype. Starting with the same feature descriptions, an inference network was constructed for the BMS system to reason about plan view images of the 10 Naval Classes. A portion of the network is shown in Figure 3. Because the nodes can represent multi-valued variables, the evidential interactions among the features can be specified directly in a manner that is intuitively meaningful. The result is a more compact and more easily understood model. This network required only 36 nodes and 35 links, as compared to the 181 nodes and 297 links used in the original version. Since the links in the network point from cause to effect, the conditional probabilities for the links do not depend on the proportion of ships of each type. This means that a spreadsheet database is no longer needed to manage changes in the model parameters.

When tested on the 52 plan view images, the BMS version produced results nearly identical to those obtained with the PROSPECTOR version. The correct class was ranked first on exactly the same set of images (39 out of 52). In fact, the two versions assigned slightly different rankings to the correct class on just 4 occasions. Overall, the average rank assigned to the correct class was the same for both systems. This is not too surprising, given that the PROSPECTOR version was supplied with a consistent set of probabilities and the network was really a tree. Under those circumstances, the PROSPECTOR method complies with the axioms of probability and the weight of diagnostic evidence is properly distributed.

One interesting implementation issue that emerged from this exercise relates to the order in which nodes are updated. Any sequential implementation of the BMS computation has to keep track of which nodes need to be activated for belief revision. Our original implementation used a stack for this purpose, and it worked well as long as only one piece of evidence was offered to the network at a time. When several perturbations were made at the same time, however, the efficiency of the propagation scheme deteriorated. There are two reasons for this. First, using a stack causes the system to bring small parts of the network into equilibrium before considering the effects of other perturbed nodes.

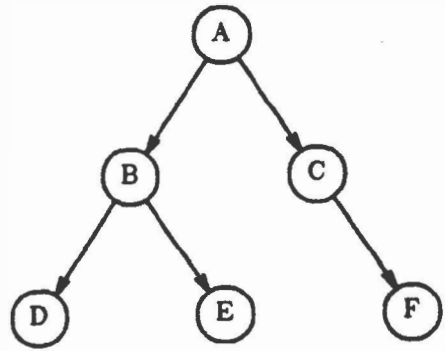

Figure 4

For example, if nodes E and F in Figure 4 are both activated, the impact from E does not propagate to C until A, B and E are in equilibrium. Achieving equilibrium here is a waste of time, however, since the effects from F will disrupt it. This can be avoided by processing activated nodes in a first-in, first-out (FIFO) order, which is more in keeping with the distributed processing spirit of the BMS computation. Second, the number of updates needed to reach equilibrium can be substantially reduced by avoiding duplicate

35

entries in the list. If a node already on the list is moved to the end whenever it receives another activation signal, updating that node is postponed until the information from its neighbors is more complete. These alternatives have been tested on a simple network consisting of 24 nodes and 23 links. Given 8 pieces of evidence simultaneously, the stack implementation reached equilibrium after 195 node updates, the FIFO version needed 108 updates, and the FIFO version with duplicates removed needed only 71 updates.

## CONCLUSIONS

For our purposes, the BMS updating method provides a flexibility, robustness and conceptual clarity that was not available with the PROSPECTOR approach. It has the additional advantage of being amenable to a straightforward hardware implementation, an important consideration in a realtime application. There was no significant difference in the performance of the two versions of the classification system on a simple task reasoning from evidence to hypothesis. However, the BMS version was much easier to understand and maintain.

### Acknowledgements

† This research would not have been possible without the cooperation of NRL's image interpretation experts.

## REFERENCES

[1] Booker, L. B. An Artificial Intelligence (AI) Approach to Ship Classification. In *Intelligent Systems: Their Development and Application*. Proceedings of the 24th Annual Technical Symposium, Washington D.C. Chapter of the ACM. Gaithersburg, MD., June, 1985, p. 29-35.

[2] Duda, R.O., Hart, P.E., and Nilsson, N.J. Subjective Bayesian Methods for Rule-Based Inference Systems. Technical Note 124, SRI International, Menlo Park, CA, January 1976.

[3] Duda, R.O., Hart, P.E., Konolige, K., and Reboh, R. A Computer-Based Consultant for Mineral Exploration. Final Report, SRI Project 6415, SRI International, Menlo Park, CA, September 1979.

[4] Kim, J. and Pearl, J., A Computational Model for Combined Causal and Diagnostic Reasoning in Inference Systems, *Proceedings of IJCAI-83*, Los Angeles, CA., August, 1985, p. 190-193.

[5] Pearl, J., How To Do With Probabilities What People Say You Can't. Technical Report CSD-850031, Computer Science Department, University of California, Los Angeles, CA, September 1985.

[6] Pearl, J., Distributed Diagnosis in Causal Models with Continuous Variables. Technical Report CSD-860051, Computer Science Department, University of California, Los Angeles, CA, December 1985.

[7] Pearl, J., On Evidential Reasoning in a Hierarchy of Hypotheses. *Artificial Intelligence* , Vol. 28, No. 1 (1986), p. 9-15.

[8] Quinlan, J., Inferno: A Cautious Approach to Uncertain Inference. *The Computer Journal,* Vol. 26, No. 3 (1983), p. 255-269

[9] Slagle, J., Gaynor, M., and Halpern, E., An Intelligent Control Strategy for Computer Consultation. *IEEE Trans. Pattern Anal. Machine Intell.* PAMI-6, (March 1984), p. 129-136.

[10] Slagle, J. and Hamburger, H. An Expert System for a Resource Allocation Problem. *Communications of the ACM,* Vol. 28, No. 9 (Sept. 1985), p. 994-1004.